\documentclass[11pt,a4paper]{article}
\usepackage[hyperref]{acl2021}
\usepackage{times}
\usepackage{latexsym}

\usepackage{microtype}

\aclfinalcopy 

\usepackage{xspace}
\usepackage{multirow}
\usepackage{xcolor,colortbl}
\usepackage{graphicx}
\usepackage{nicematrix}
\usepackage{booktabs}
\usepackage{bm}
\usepackage{subfig}
\usepackage{commath}
\usepackage{adjustbox}
\usepackage{fbox}
\usepackage[subpreambles=true]{standalone}

\newcommand{\tapas}{\textsc{TAPAS}\xspace}
\newcommand{\tabfact}{\textsc{TabFact}\xspace}
\newcommand{\semtabfact}{\textsc{SEM-TAB-FACTS}\xspace}
\newcommand{\bert}{\textsc{BERT}\xspace}
\definecolor{myblue}{HTML}{DAE8FC}
\newcommand{\highlightcell}{\cellcolor{myblue}}
\newcommand{\err}[1]{\textsubscript{~$\pm$#1}}

\title{Volta at SemEval-2021 Task 9: Statement Verification and Evidence Finding with Tables using TAPAS and Transfer Learning}

\author{Devansh Gautam\thanks{\enspace The authors have contributed equally.} \qquad Kshitij Gupta\footnotemark[1] \qquad Manish Shrivastava \\
  International Institute of Information Technology Hyderabad \\
  \texttt{\{devansh.gautam,kshitij.gupta\}@research.iiit.ac.in},\\ \texttt{m.shrivastava@iiit.ac.in}
  }

\date{}

\begin{document}
\maketitle
\begin{abstract}
    Tables are widely used in various kinds of documents to present information concisely. Understanding tables is a challenging problem that requires an understanding of language and table structure, along with numerical and logical reasoning. In this paper, we present our systems to solve Task 9 of SemEval-2021: \textit{Statement Verification and Evidence Finding with Tables}~(\semtabfact). The task consists of two subtasks: (A) Given a table and a statement, predicting whether the table supports the statement and (B) Predicting which cells in the table provide evidence for/against the statement. We fine-tune \tapas (a model which extends \bert's architecture to capture tabular structure) for both the subtasks as it has shown state-of-the-art performance in various table understanding tasks. In subtask A, we evaluate how transfer learning and standardizing tables to have a single header row improves \tapas' performance. In subtask B, we evaluate how different fine-tuning strategies can improve \tapas' performance. Our systems achieve an F1 score of 67.34 in subtask A three-way classification, 72.89 in subtask A two-way classification, and 62.95 in subtask B.
\end{abstract}

\section{Introduction}

There has been extensive work on verifying if a given textual context supports a given statement. Even though tables are also widely used to convey information, especially in scientific texts, there has been comparatively less work on verifying if a given table supports a statement. To this end, SemEval 2021 Task 9~\citep{wang-etal-2021-semeval} focuses on statement verification and evidence finding for tables from scientific articles in the English language. The task is divided into two subtasks - A and B. The aim of subtask A is to classify whether a given statement is entailed or refuted according to the given table and associated table metadata~(such as captions and legends) or whether the statement's truth is unknown as it cannot be determined from the table. The aim of subtask B is to classify each cell in the table as relevant or irrelevant in determining whether the statement is entailed or refuted from the tabular evidence (the truth value of the statement is also provided).

Our systems use \tapas~\citep{herzig-etal-2020-tapas} trained with intermediate pre-training ~\citep{eisenschlos-etal-2020-understanding} for both the subtasks. For subtask A, we fine-tune \tapas after adding a three-way classification head on top for classifying the statement as entailed/refuted/unknown. We also evaluate how transfer learning and standardizing tables to have a single header row can improve \tapas' performance. Due to the similarity between subtask B and table question-answering~(which involves cell selection or cell selection followed by aggregation), we use the \tapas architecture previously used for table question-answering and fine-tune it to select the relevant cells. We also evaluate how different fine-tuning strategies can improve \tapas' performance on evidence finding.

Our systems achieve an F1-micro score of 67.34 in subtask A and 72.89 in subtask A if the unknown statements are not considered while calculating the metrics~(however, classifying entailed/refuted statements as unknown is still penalized). Our submitted system achieves an F1 score of 62.95 in subtask B. During the post-evaluation phase, we modified our system and achieved an F1-score of 65.48 in subtask B.

The code for our systems is available at \url{https://github.com/devanshg27/sem-tab-fact}.
\section{Background}

\begin{figure*}[!t]
\centering
\resizebox{\columnwidth}{!}{%
\begin{NiceTabular}{lcccc}[hvlines]

\Block{3-1}{} & \Block{1-1}{(1)} & \Block{1-1}{(2)} & \Block{1-1}{(3)} & \Block{1-1}{(4)} \\ 
 & \Block{1-2}{English Language} & & \highlightcell \Block{1-2}{ English} & \highlightcell\\  
 &  Frequency & Percent & Frequency & \highlightcell Percent \\ 
\highlightcell AQA & 241539 & 61.6 & 84742 & \highlightcell 55.7 \\ 
WJEC & 83219 & 21.2 & 39650 & \highlightcell 26.1 \\ 
Pearson & 37194 & 9.5 & 18815 & \highlightcell 12.4 \\ 
OCR & 30061 & 7.7 & 8818 & \highlightcell 5.8 \\ 
Total & 392015 & & 152025 & \\ 
\end{NiceTabular}%
}
\begin{tabular}{|p{\columnwidth}|}
\hline
\small \textbf{Caption:} Number of GCSE Full Course entries by Awarding Body (KS4 Results tables, 2014) \\
\small \textbf{Legend:} Note. Number of GCSE Full Course entries in the summer season of the academic year 2012-2013. AQA (The Assessment and Qualifications Alliance); WJEC (Welsh Joint Education Committee); OCR (Oxford, Cambridge and RSA Examinations); CCEA (Council for the Curriculum, Examinations and Assessment). We do not show the information of an additional awarding body that accounts for almost no entries.\\
\hline
\end{tabular}\\
\vspace{8pt}
\begin{tabular}{|ll|}
\hline
{\small \emph{Entailed:}} &{\small 1. The highest Frequency, not counting the Total, is 84742.}\\
                          &{\small 2. The highest English Percent is for AQA}\\

{\small \emph{Refuted:}}  &{\small 1. The highest Percent value for OCR is 5.8 }\\
                          &{\small 2. The lowest total is 392015}\\

{\small \emph{Unknown:}} & {\small 1. First, this is due to technical problems in providing Unique}\\
                         & {\small \hspace{0.85em} Candidate Numbers (UPN) for all candidates.}\\
                         & {\small 2. This is for four main reasons.}\\
\hline
\end{tabular}%
\caption{An example from the \semtabfact dataset: Table \texttt{A1} From \texttt{10262.xml} along with its caption and legend. Some example statements of each class associated with this table are also shown. The highlighted cells are the relevant cells for entailed statement 2.}
\label{fig:example}
\end{figure*}

Verifying if the given textual evidence supports a given statement is a fundamental natural language processing problem. It has been extensively studied under different tasks such as RTE (Recognizing Textual Entailment)~\citep{10.1007/11736790_9}, NLI (Natural Language Inference)~\citep{bowman-etal-2015-large}, FEVER (Fact Extraction and VERification)~\citep{thorne-etal-2018-fever}. In recent years, large-scale pre-trained models~\citep{devlin-etal-2019-bert, peters-etal-2018-deep, NEURIPS2019_dc6a7e65, liu2019roberta} have dominated these tasks and have achieved close-to-human performance. NLVR~\citep{suhr-etal-2017-corpus} and NLVR2~\citep{suhr-etal-2019-corpus} focus on verifying a statement given an image as evidence. INFOTABS~\citep{gupta-etal-2020-infotabs} and \tabfact~\citep{Chen2020TabFact:} focus on verifying a statement given a table from Wikipedia\footnote{\url{https://www.wikipedia.org/}} as evidence.

\citet{neeraja-etal-2021-incorporating} propose simple modifications to how information is presented to existing textual models such as RoBERTa~\citep{liu2019roberta} to improve tabular fact verification. Along with releasing \tabfact, \citet{Chen2020TabFact:} also discuss two promising approaches for tabular fact verification, Latent Program Algorithm (LPA) and Table-BERT. LPA is a semantic parsing approach that parses statements into programs (logical forms) and executes the programs against the table to predict the entailment decision. Most of the current models~\citep{zhong-etal-2020-logicalfactchecker, shi-etal-2020-learn, yang-etal-2020-program} for \tabfact are semantic parsing approaches similar to LPA. Table-BERT encodes the linearized tables and statements using \bert-based models and directly predicts the entailment decision. \citet{zhang-etal-2020-table} inject table structural information into the mask of the self-attention layer of \bert-based models, which helps the model learn better table representations. \tapas~\citep{herzig-etal-2020-tapas} extends \bert's architecture to capture the tabular structure, and it showed competitive performance on various table question answering datasets: SQA~\citep{iyyer-etal-2017-search}, WTQ~\citep{pasupat-liang-2015-compositional} and WikiSQL~\citep{zhong2017seq2sql}. \citet{eisenschlos-etal-2020-understanding} add an intermediate pre-training step before the fine-tuning step to \tapas and show that it achieves state-of-the-art results on \tabfact and SQA~\citep{iyyer-etal-2017-search}. Their model is still 8 points behind human performance on \tabfact since tabular fact verification involves table understanding and complex reasoning.

While \tabfact also focuses on fact verification using tables as evidence, it focuses on tables from Wikipedia, whereas SemEval-2021 Task 9 (\semtabfact) instead focuses on tables from scientific articles and has a subtask related to evidence finding. Also, \tabfact did not have a neutral/unknown class, which they left out because of low inter-worker agreement due to confusion with refuted class. Figure~\ref{fig:example} shows an example of a table from the \semtabfact dataset and the labels for the two subtasks.
\section{System Overview}

\begin{table*}[htbp]
\resizebox{\textwidth}{!}{%
\begin{tabular}{lccc}
\toprule
 & \textbf{Train~(Auto) Set} & \textbf{Train~(Manual) Set} & \textbf{Dev Set} \\ \midrule
Total number of tables with \texttt{<thead>} tag & 1977 & 980 & 52 \\
Number of tables with correct header prediction & 1855(93.83\%) & 918(93.67\%) & 51(98.08\%) \\
Number of tables with header prediction error is $\le1$ & 1966(99.44\%) & 972(99.18\%) & 52(100\%) \\
\bottomrule
\end{tabular}%
}
\caption{Header Prediction Statistics}
\label{tab:header-table}
\end{table*}

In this section, we provide a general overview of our systems for the two subtasks. We use \tapas for both subtasks.

\subsection{Subtask A: Statement Verification} \label{sec:overview-subtask-a}

\paragraph{Pre-processing} Since \tapas only works on tables with single cells~(cells which do not span multiple columns/rows) only, we first convert the tables with multi-row/multi-column cells to tables with only single cells by duplicating the value of the cell in every single cell the multi-row/multi-column cell spans. An example of the pre-processing is shown in Figure~\ref{fig:fig:header1}.

\begin{figure}%
\centering
    \subfloat[Converting multi-row/multi-column cells to single cells\label{fig:fig:header1}]{\resizebox{\columnwidth}{!}{%
\centering
\begin{tabular}{|l|c|c|c|c|}
\hline
 & \multicolumn{1}{l|}{(1)} & \multicolumn{1}{l|}{(2)} & \multicolumn{1}{l|}{(3)} & \multicolumn{1}{l|}{(4)} \\ \hline
 & \multicolumn{1}{l|}{English Language} & \multicolumn{1}{l|}{English Language} & \multicolumn{1}{l|}{English} & \multicolumn{1}{l|}{English} \\ \hline
 & \multicolumn{1}{l|}{Frequency} & \multicolumn{1}{l|}{Percent} & \multicolumn{1}{l|}{Frequency} & \multicolumn{1}{l|}{Percent} \\ \hline
AQA & 241539 & 61.6 & 84742 & 55.7 \\ \hline
WJEC & 83219 & 21.2 & 39650 & 26.1 \\ \hline
Pearson & 37194 & 9.5 & 18815 & 12.4 \\ \hline
OCR & 30061 & 7.7 & 8818 & 5.8 \\ \hline
Total & 392015 &  & 152025 &  \\ \hline
\end{tabular}%
}}%
\\
    \subfloat[Standardizing the header rows of the table with single cells \label{fig:fig:header2}]{\resizebox{\columnwidth}{!}{%
\centering
\begin{tabular}{|l|c|c|c|c|}
\hline
 & \begin{tabular}[c]{@{}l@{}}(1)\\ English Language\\ Frequency\end{tabular} & \begin{tabular}[c]{@{}l@{}}(2)\\ English Language\\ Percent\end{tabular} & \begin{tabular}[c]{@{}l@{}}(3)\\ English\\ Frequency\end{tabular} & \begin{tabular}[c]{@{}l@{}}(4)\\ English\\ Percent\end{tabular} \\ \hline
AQA & 241539 & 61.6 & 84742 & 55.7 \\ \hline
WJEC & 83219 & 21.2 & 39650 & 26.1 \\ \hline
Pearson & 37194 & 9.5 & 18815 & 12.4 \\ \hline
OCR & 30061 & 7.7 & 8818 & 5.8 \\ \hline
Total & 392015 &  & 152025 &  \\ \hline
\end{tabular}%
}}%
    \caption{Pre-processing and header standardization applied to the table shown in Figure~\ref{fig:example}.}%
    \label{fig:header}%
\end{figure}

\paragraph{Header Standardization} We experiment with standardizing the pre-processed tables with multi-row headers to tables with a single header row since \tapas was pre-trained on single header tables and \tabfact~(which we want to use for transfer learning) also contains single header tables. We first predict the number of header rows using the following rules:

\begin{enumerate}
    \item In many pre-processed tables, we found that the left-most column contained row names, and either (a) all the header cells in the left-most column were empty, or (b) the cell value at the top-left corner was repeated in all the header cells below it, or (c) the cell at the top-left corner was not empty, but the header cells below it were empty. Based on these cases, we initially estimate the number of header rows as the number of rows at the top, such that all cells in the left-most column in those rows are either empty or have the same value as the cell at the top-left corner.
    \item We also found that in many cases, there were multi-column cells in the header, which had more specific sub-headers in the rows below. To handle these cases, we increment the estimate of header rows until no two adjacent columns have the same header cell values.
\end{enumerate}

We merge the predicted header rows into a single row by joining each column's header cell values into a single cell with a newline as a separator. An example of header standardization is shown in Figure~\ref{fig:fig:header2}. We were provided with HTML versions of the tables in the training and development set. We compare our predictions against the \texttt{<thead>} tags in the HTML tables to analyze our header prediction system's performance. The results are shown in Table~\ref{tab:header-table}. We also find that in almost all of the cases, the predictions are either correct or have an error of $\pm1$.

To study the effect of header standardization, we will train all our systems with and without header standardization.

\begin{figure*}[htbp]%
\centering
    \subfloat[\centering Subtask A\label{fig:modela}]{{\includegraphics[width=6cm]{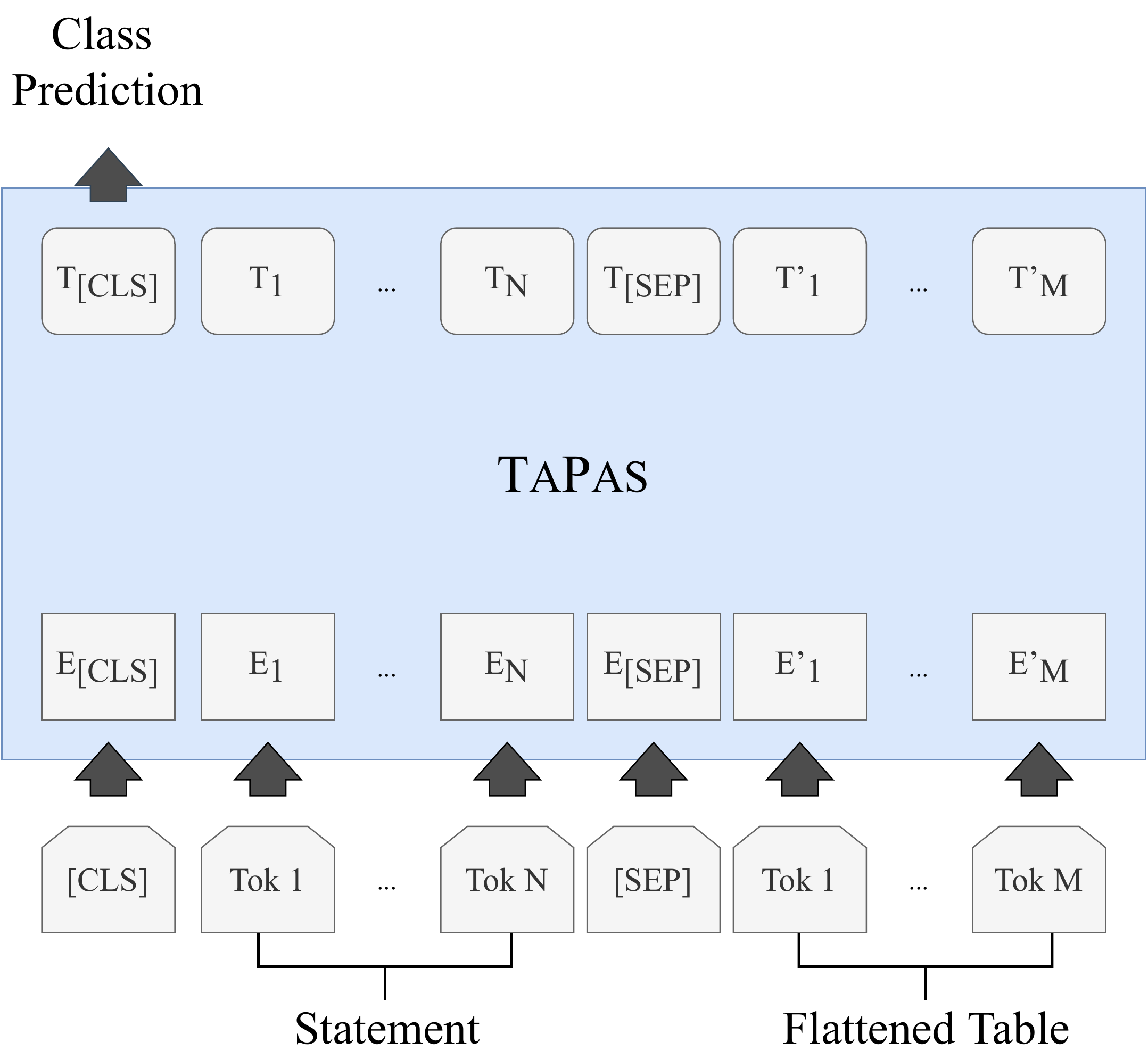} }}%
    \qquad
    \qquad
    \subfloat[\centering Subtask B\label{fig:modelb}]{{\includegraphics[width=6cm]{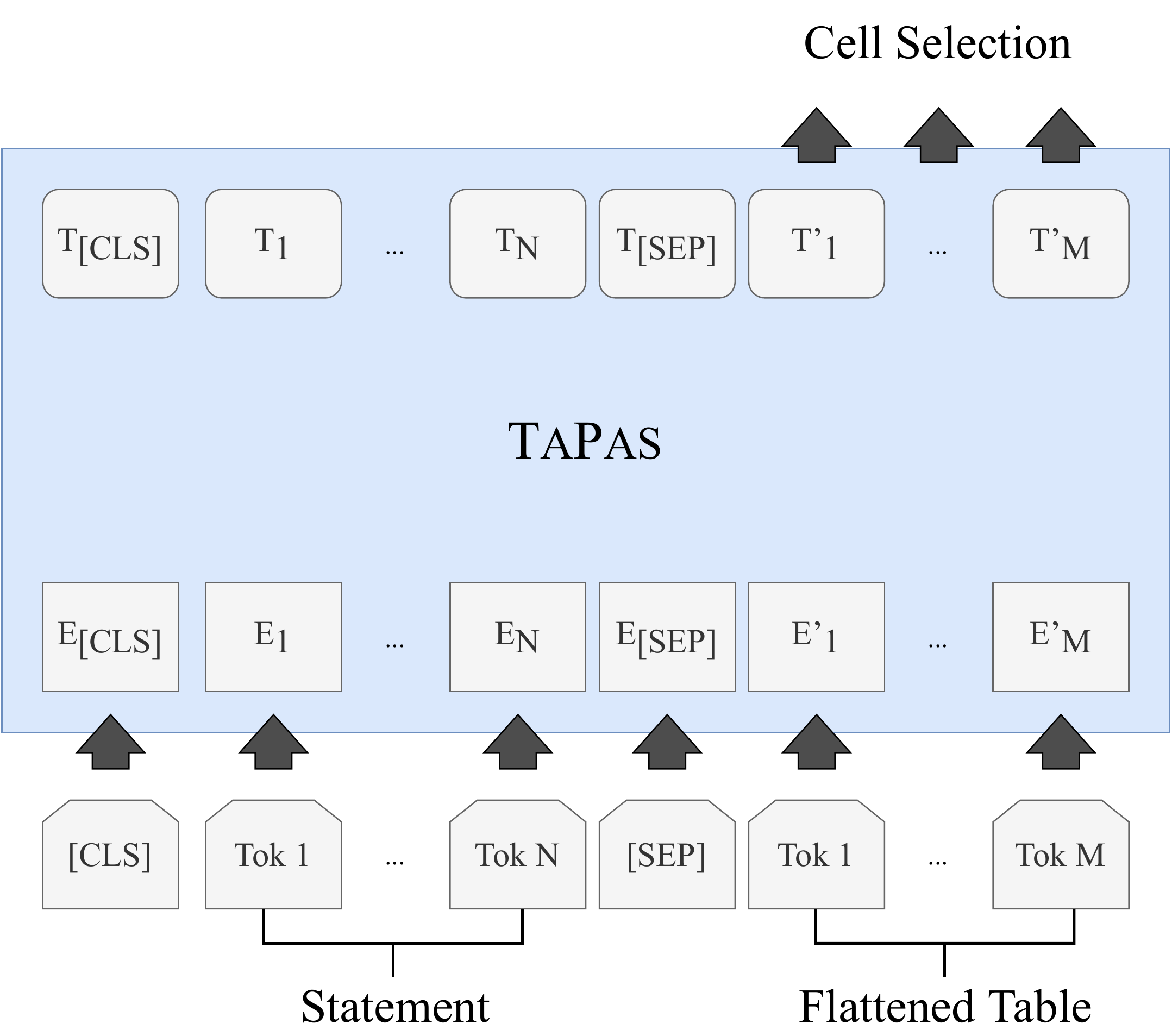} }}%
    \caption{The architecture of our models}%
    \label{fig:model}%
\end{figure*}

\paragraph{Model} Our model takes the following input: \texttt{[CLS] <statement> [SEP] <flattened table>}, which is tokenized using the standard \bert tokenizer. We compute the class probabilities using a linear layer with a softmax activation function on top of the output of the \texttt{[CLS]} token, as shown in Figure~\ref{fig:modela}. We use the weighted cross-entropy loss, which helps in handling imbalance in the class sizes:

$$H_y(y')=-\sum_{i}\sum_{k=1}^{K}w_{k}y_{ik} \cdot \log(y'_{ik})$$

Where $y_{ik}$ denotes the ground truth label, it is $1$ if $k$ is the true class label of the $i^{th}$ token, and $0$ otherwise, $y'_{ik}$ is the corresponding model probability prediction and $w_k$ is the weight for class $k$. We set $w_k$ as the size of the biggest class divided by the size of class $k$.

To analyze how transfer learning can improve performance, we compare the following approaches:

\begin{itemize}
    \item \textbf{\tapas-stf:} We use the publicly available \tapas checkpoint which has been pre-trained with a masked language modeling objective and fine-tune it on the \semtabfact dataset provided by the task organizers.
    \item \textbf{\tapas-tf:} As a baseline, we directly use the publicly available \tapas checkpoint, which had been fine-tuned on \tabfact without any further fine-tuning on \semtabfact. Since \tabfact has only entailed/refuted labels, this model is a binary classifier and does not predict the unknown class's probabilities.
    \item \textbf{\tapas-tf-stf:} We use the publicly available \tapas checkpoint, which had been fine-tuned on \tabfact and further fine-tune it on the \semtabfact dataset released by the task organizers. This is our submitted model for subtask A.
\end{itemize}

\subsection{Subtask B: Evidence Finding}

\paragraph{Pre-processing and Header Standardization} We convert the multi-row/multi-column cells and standardize the header rows as discussed in Section~\ref{sec:overview-subtask-a}. The relevant/irrelevant labels of the multi-row/multi-column cells are duplicated to all the single cells they span. We consider the relevant/irrelevant labels only for the cells of the non-header rows as \tapas does not make predictions for header cells. Based on the performance of header standardization in subtask A~(which we will discuss in Section~\ref{sec:results}), we standardize headers for all our models in this subtask.

\paragraph{Model} Our model takes the following input: \texttt{[CLS] <statement> [SEP] <flattened table>}, which is tokenized using the standard \bert tokenizer. We show the architecture of our model in Figure~\ref{fig:modelb}. Our model computes token-level logits using a linear layer on top of each token's last hidden state output, which are used to compute cell-level logits by averaging the logits of the tokens in each cell. The probability of selection for each cell is calculated from the cell-level logits using the sigmoid function. We use the weighted binary cross-entropy loss which helps in handling class imbalance:

$$
\resizebox{\columnwidth}{!}{$\displaystyle
H_y(y')=-\sum_{i}w_{p}y_{i} \cdot \log y'_{i}+\left(1-y_{i}\right) \cdot \log \left(1-y'_{i}\right)
$}
$$

Where $y_{i}$ denotes the ground-truth label, it is $1$ if the $i^{th}$ token is part of any relevant cell, and $0$ otherwise, $y'_{i}$ is the corresponding model probability prediction, and $w_p$ denotes the weight of the positive~(relevant) class. We set $w_p$ to $10$.

\begin{table*}[!htbp]
\centering
\subfloat[Subtask A\label{tab:taska-data}]{
\resizebox{\textwidth}{!}{%
\begin{tabular}{lcccc}
\toprule
 & \textbf{\#Tables} & \textbf{\#Entailed statements} & \textbf{\#Refuted statements} & \textbf{\#Unknown statements} \\ \midrule
\textbf{Train (Auto-generated)} & 1980 & 92136 & 87209 & 0 \\
\textbf{Train (Manually annotated)} & 981 & 2818 & 1688 & 0 \\
\textbf{Train (with unknown statements)} & 981 & 2818 & 1688 & 4506 \\
\textbf{Validation} & 52 & 250 & 213 & 93 \\
\textbf{Test} & 52 & 274 & 248 & 131 \\ \bottomrule
\end{tabular}%
}}
\\
\subfloat[Subtask B\label{tab:taskb-data}]{
\resizebox{\textwidth}{!}{%
\begin{tabular}{lccccc}
\toprule
 & \textbf{\#Tables} & \textbf{\#Entailed statements} & \textbf{\#Refuted statements} & \textbf{\#Relevant cells} & \textbf{\#Irrelevant cells} \\ \midrule
\textbf{Train (auto-generated)} & 1980 & 92136 & 87209 & 1039058 & 15467957 \\
\textbf{Validation} & 51 & 233 & 191 & 3048 & 28495 \\
\textbf{Test} & 52 & 251 & 219 & 3458 & 26724 \\ \bottomrule
\end{tabular}%
}}
\caption{Dataset Statistics for each subtask}
\label{tab:task-data}
\end{table*}

Due to the similarity of evidence finding with table question-answering, we use the publicly available \tapas checkpoint, which was fine-tuned in a chain on SQA, WikiSQL, and finally WTQ. We compare the following fine-tuning strategies:

\begin{itemize}
    \item \textbf{WTQ-base:} As a baseline, we fine-tune our model directly for relevant cell selection on \semtabfact.
    \item \textbf{WTQ-statement:} We again fine-tune the model for relevant cell selection on \semtabfact, but we try to include the information on whether the statement was entailed/refuted by modelling the statement as `\textit{Which cells entail ``{\textless}statement{\textgreater}"?}' or `\textit{Which cells refute ``{\textless}statement{\textgreater}"?}'. {\textless}statement{\textgreater} denotes the original statement.
    \item \textbf{WTQ-separate:} We fine-tune two separate models, one which predicts the relevant cells for entailed statements and another one for refuted statements. This is our submitted system for subtask B.
\end{itemize}

During the post-evaluation phase, we experimented with the publicly available \tapas checkpoint, which was fine-tuned on \tabfact. Similar to the systems described above, we compare three systems based on this checkpoint: TABFACT-base, TABFACT-statement, and TABFACT-separate.

\paragraph{Post-Processing} We further apply post-processing steps to obtain the final prediction from the cell classification. To predict the header's relevant cells, we select the header cells for any column with cells selected as a relevant cell. We label multi-row/multi-column cells as relevant if any of the single cells they span are predicted as relevant.
\section{Experimental Setup}

\subsection{Data Description}

We used the dataset provided by the task organizers for both subtasks. We did not use the table metadata in our systems.

For subtask A, dataset statistics and the official splits are shown in Table~\ref{tab:taska-data}. The provided training sets do not have any statements of the unknown class. So, we used the manually annotated training set to create a training set with unknown statements. Each statement of the manually annotated training set was added as an unknown statement to a different table chosen randomly. We used this dataset for training all our models for subtask A.

For subtask B, dataset statistics and the official splits are shown in Table~\ref{tab:taskb-data}. We use the auto-generated training set for training all our models in subtask B.

\begin{table*}[!t]
\centering
\resizebox{\textwidth}{!}{
\begin{tabular}{lcccccc}
\toprule
\multirow{2}{*}[-2.5pt]{\textbf{F1 Score}} & \multicolumn{3}{c}{\textbf{Validation Set}} & \multicolumn{3}{c}{\textbf{Test Set}}\\
\cmidrule(lr){2-4}
\cmidrule(lr){5-7}
  & \textbf{\tapas-stf} & \textbf{\tapas-tf} & \textbf{\tapas-tf-stf} & \textbf{\tapas-stf} & \textbf{\tapas-tf} & \textbf{\tapas-tf-stf}\\ \midrule
\multicolumn{7}{l}{\textit{Without header standardization}} \\ \midrule
2-way micro & $\bm{72.1} \err{0.43}$ & $69.42$ & $71.01 \err{0.99}$ & $68.01 \err{0.28}$ & $70.97$ & $\bm{72.97} \err{1.37}$\\
3-way micro & $\bm{66.41} \err{0.48}$ & $58.97$ & $65.76 \err{0.37}$ & $61.59 \err{0.02}$ & $57$ & $\bm{65.15} \err{0.81}$\\
Refuted & $67.95 \err{0.98}$ & $64.31$ & $\bm{70.32} \err{0.91}$ & $62.04 \err{0.45}$ & $64.05$ & $\bm{69.13} \err{0.74}$\\
Entailed & $67.8 \err{0.36}$ & $58.94$ & $\bm{68.09} \err{1.24}$ & $64.89 \err{0.49}$ & $61.9$ & $\bm{67.23} \err{1.18}$\\
Unknown & $\bm{49.76} \err{0.73}$ & $0$ & $47.52 \err{3.52}$ & $\bm{47.58} \err{0.8}$ & $0$ & $46.43 \err{1.88}$\\ \midrule
\multicolumn{7}{l}{\textit{With header standardization}} \\ \midrule
2-way micro & $71.34 \err{0.96}$ & $72.78$ & $\bm{74.35} \err{1.14}$ & $68.67 \err{0.9}$ & $73.79$ & $\bm{73.87} \err{0.87}$\\
3-way micro & $66.16 \err{0.64}$ & $61.11$ & $\bm{69.16} \err{0.58}$ & $61.99 \err{0.8}$ & $59.32$ & $\bm{66.95} \err{0.27}$\\
Refuted & $68.22 \err{0.29}$ & $65.98$ & $\bm{73.2} \err{0.83}$ & $61.42 \err{1.9}$ & $65.7$ & $\bm{70.39} \err{0.44}$\\
Entailed & $67.98 \err{0.43}$ & $63.67$ & $\bm{70} \err{1.69}$ & $65.67 \err{0.21}$ & $65.38$ & $\bm{68.9} \err{0.48}$\\
Unknown & $49.9 \err{3.07}$ & $0$ & $\bm{50.91} \err{3.99}$ & $48.27 \err{1.55}$ & $0$ & $\bm{50.89} \err{3.93}$\\ \bottomrule
\end{tabular}%
}
\caption{Performance on subtask A: Mean and standard deviation of the metrics from 3 independent runs. In the case of \tapas-tf, we calculate the metrics using the publicly available \tapas checkpoint fine-tuned on \tabfact.}
\label{tab:resultsA}
\end{table*}

\begin{table*}[!t]
\centering
\resizebox{\textwidth}{!}{
\begin{tabular}{lcccccc}
\toprule
\multirow{2}{*}[-2.5pt]{\textbf{Model}} & \multicolumn{3}{c}{\textbf{Validation Set}} & \multicolumn{3}{c}{\textbf{Test Set}}\\
\cmidrule(lr){2-4}
\cmidrule(lr){5-7}
  & \textbf{F1} & \textbf{F1\textsubscript{entailed}} & \textbf{F1\textsubscript{refuted}} & \textbf{F1} & \textbf{F1\textsubscript{entailed}} & \textbf{F1\textsubscript{refuted}}\\ \midrule
WTQ-base & $55.39 \err{0.53}$ & $64.07 \err{0.65}$ & $48.66 \err{0.47}$ & $61.36 \err{1.47}$ & $68.47 \err{2.49}$ & $\bm{52.75} \err{1.15}$\\
WTQ-statement & $55.18 \err{1.78}$ & $63.36 \err{3.16}$ & $48.45 \err{0.8}$ & $58.93 \err{2.49}$ & $65.22 \err{4.38}$ & $51.27 \err{0.54}$\\
WTQ-separate & $\bm{56.46} \err{0.43}$ & $\bm{66.91} \err{0.3}$ & $\bm{48.74} \err{1.01}$ & $\bm{62.26} \err{0.79}$ & $\bm{71.87} \err{1.2}$ & $50.79 \err{1.86}$\\ \midrule
\multicolumn{7}{l}{\textit{During Post-Evaluation Phase}} \\ \midrule
TABFACT-base & $58.41 \err{0.84}$ & $64.88 \err{1.37}$ & $54.02 \err{0.91}$ & $61.46 \err{0.33}$ & $67.32 \err{1.01}$ & $54.47 \err{0.55}$\\
TABFACT-statement & $58.92 \err{1.69}$ & $65.41 \err{1.95}$ & $\bm{54.18} \err{1.69}$ & $62.78 \err{1.71}$ & $68.44 \err{2.34}$ & $\bm{55.8} \err{1.36}$\\
TABFACT-separate & $\bm{59.47} \err{0.23}$ & $\bm{68.06} \err{0.79}$ & $53.16 \err{1.18}$ & $\bm{65.01} \err{0.6}$ & $\bm{74.18} \err{0.6}$ & $54.48 \err{0.58}$\\ \bottomrule
\end{tabular}%
}
\caption{Performance on subtask B: Mean and standard deviation of the metrics from 3 independent runs}
\label{tab:resultsB}
\end{table*}

\subsection{Implementation}

For the implementation of our systems, we used the HuggingFace Transformers\footnote{\hspace{1pt}\includegraphics[height=6.5pt]{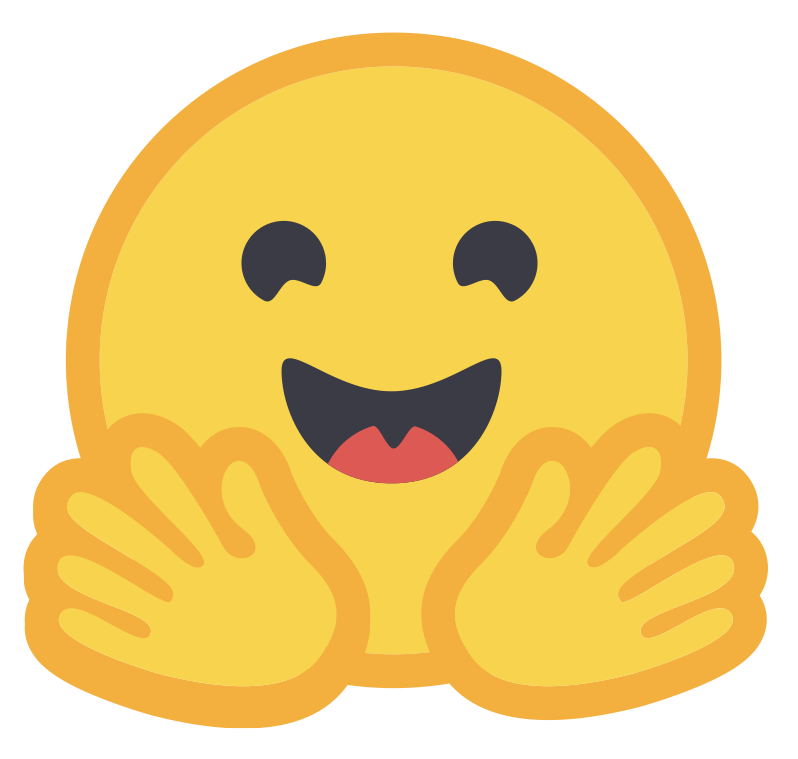} Transformers, v4.2.0, \url{https://huggingface.co/transformers/}} library\citep{wolf-etal-2020-transformers} and we used the AdamW optimizer available in PyTorch\footnote{PyTorch, v1.7.1, \url{https://pytorch.org/}}~\citep{NEURIPS2019_9015} with the default parameters (learning rates are specified below). All models were fine-tuned using a single Nvidia GeForce RTX 2080 Ti GPU.

We used the base variant of \tapas, which has a hidden dimension of 768 in all our models. All the \tapas checkpoints we used had been trained with intermediate pre-training and used relative position embeddings~(the position index reset when a new cell starts).

For subtask A, we first fine-tuned the classifier head with the \tapas layers frozen for $3$ epochs with a learning rate of $1^{-5}$ and then fine-tuned the whole model for $10$ epochs with a learning rate of $1^{-6}$. We used a batch size of $8$. We saved a checkpoint every $100$ steps and selected the best checkpoint based on the validation set performance.

For subtask B, we fine-tuned the whole model for $5000$ steps with a learning rate of $1^{-6}$. We used a batch size of $8$. We saved a checkpoint every $50$ steps and selected the best checkpoint based on the validation set performance.

\subsection{Evaluation Metrics}

In subtask A, two evaluation metrics are used. The first evaluation metric used is the standard F1-micro score for three-way classification. The second metric again calculates the F1-micro score but does not consider statements with their ground truth label as the unknown class for evaluation; however, classifying the entailed/refuted statements as unknown is penalized.

\begin{table*}[!t]
\centering
\resizebox{\textwidth}{!}{
\begin{tabular}{lcccc}
\toprule
\multirow{2}{*}{} & \multicolumn{2}{c}{\textbf{Validation Set}} & \multicolumn{2}{c}{\textbf{Test Set}}\\
\cmidrule(lr){2-3}
\cmidrule(lr){4-5}
  & \textbf{Length($\le512$)} & \textbf{Length($>512$)} & \textbf{Length($\le512$)} & \textbf{Length($>512$)}\\ \midrule
\multicolumn{5}{l}{\textit{Distribution - Number of samples}} \\ \midrule
Subtask A & $431(77.52\%)$ & $125(22.48\%)$ & $616(94.33\%)$ & $37(5.67\%)$\\
Subtask B & $345(81.37\%)$ & $79(18.63\%)$ & $442(94.04\%)$ & $28(5.96\%)$\\
\midrule
\multicolumn{5}{l}{\textit{Performance of each task's best model}} \\ \midrule
Subtask A 2-way F1-micro & $77.83\err{0.57}$ & $65.49\err{3.13}$ & $73.83\err{0.83}$ & $74.44\err{1.57}$\\
Subtask A 3-way F1-micro & $73.13\err{1.13}$ & $55.53\err{1.4}$ & $66.71\err{0.35}$ & $54.74\err{1.12}$\\
Subtask B F1 & $62.79\err{0.39}$ & $45.91\err{0.68}$ & $65.38\err{0.63}$ & $58.98\err{1.22}$\\
\bottomrule
\end{tabular}%
}
\caption{Results on long sequences}
\label{tab:longsequences}
\end{table*}

In subtask B, the evaluation metric used is the standard F1 score with relevant cells as the positive class. If multiple minimal sets of cells can be used to determine the statement's truth value, the dataset contains all of these versions. The score for that statement is calculated by comparing the prediction against each ground truth version and considering the highest score.
\section{Results} \label{sec:results}

\paragraph{Subtask A} The performance of the various systems we considered in subtask A is shown in Table~\ref{tab:resultsA}. Header standardization improves the performance of all the systems we compared. Transfer learning from \tabfact also improves the performance of our systems. Surprisingly, \tapas-tf without any fine-tuning on \semtabfact has a better two-way F1-micro score than \tapas-stf. This shows us the potential of transfer learning from \tabfact in subtask A.

From the confusion matrix shown in Figure~\ref{fig:confusion1a}, we observe that our model struggles with the unknown class and often misclassifies it as refuted.

\paragraph{Subtask B} The performance of the various systems we considered in subtask A is shown in Table~\ref{tab:resultsB}. Modifying the statement to include entailed/refuted class information leads to a small drop in performance for the models fine-tuned on question-answering earlier and led to a small increase in performance in models fine-tuned on \tabfact. Separate models for entailed/refuted statements perform the best among the systems we considered. It significantly improves the performance on entailed statements, with a little drop in performance on refuted statements. Surprisingly, we observe that transfer learning from \tabfact performs better than transfer learning from WTQ, even though it is a cell selection task. We believe this is because the model has to predict the cells that can be used as evidence for table entailment. The token-level embeddings of the model fine-tuned on \tabfact are better for this task than the model fine-tuned on WTQ, which is instead a question-answering dataset. 

\paragraph{Long Inputs} The maximum number of tokens supported by our system is $512$. In sequences longer than $512$ tokens, the tables are truncated row by row to fit in $512$ tokens. We compare our system's performance on these long sequences and sequences that fit within $512$ tokens. The results are shown in Table~\ref{tab:longsequences}. We find a significant drop in performance on sequences longer than $512$ tokens which had to be truncated.

\begin{figure}
    \centering
    \subfloat[\centering Subtask A\label{fig:confusion1a}]{\resizebox{0.6\columnwidth}{!}{
  \marginbox{-1em 0em 1em 0em}{
\def\myConfMat{{
{  186,  72,  21},  
{   50, 195,  52},  
{   12,   7,  58},  
}}

\def\classNames{{"R","E","U"}} 

\def\numClasses{3} 

\def\myScale{1.25} 
\begin{tikzpicture}[
    scale = \myScale,
    ]

\tikzset{vertical label/.style={rotate=90,anchor=east}}   
\tikzset{diagonal label/.style={rotate=45,anchor=north east}}

\foreach \y in {1,...,\numClasses} 
{
    \node [anchor=east] at (0.4,-\y) {\pgfmathparse{\classNames[\y-1]}\pgfmathresult}; 
    
    \foreach \x in {1,...,\numClasses}  
    {
    \def\totSamples{0}
    \foreach \ll in {1,...,\numClasses}
    {
        \pgfmathparse{\myConfMat[\ll-1][\x-1]}   
        \xdef\totSamples{\totSamples+\pgfmathresult} 
    }
    \pgfmathparse{\totSamples} \xdef\totSamples{\pgfmathresult}  
    
    \begin{scope}[shift={(\x,-\y)}]
        \def\mVal{\myConfMat[\y-1][\x-1]} 
        \pgfmathtruncatemacro{\r}{\mVal}   %
        \pgfmathtruncatemacro{\p}{round(\r/\totSamples*100)}
        \coordinate (C) at (0,0);
        \ifthenelse{\p<50}{\def\txtcol{black}}{\def\txtcol{white}} 
        \node[
            draw,                 
            text=\txtcol,         
            align=center,         
            fill=black!\p,        
            minimum size=\myScale*10mm,    
            inner sep=0,          
            ] (C) {\r\\\p\%};     
        \ifthenelse{\y=\numClasses}{
        \node [] at ($(C)-(0,0.75)$) 
        {\pgfmathparse{\classNames[\x-1]}\pgfmathresult};}{}
    \end{scope}
    }
}
\coordinate (yaxis) at (-0.3,0.5-\numClasses/2);  
\coordinate (xaxis) at (0.5+\numClasses/2, -\numClasses-1.25); 
\node [vertical label] at (yaxis) {Predicted Class};
\node []               at (xaxis) {Target Class};
\end{tikzpicture}}}}%
\subfloat[\centering Subtask B]{\resizebox{0.4\columnwidth}{!}{
  \marginbox{-2em 0em 2em 0em}{
\def\myConfMat{{
{  2460,  90},  
{   213, 256},  
}}

\def\myConfMatRatio{{
{  24596,  903},  
{   2128, 2555},  
}}

\def\classNames{{"I","R"}} 

\def\numClasses{2} 

\def\myScale{1.25} 
\begin{tikzpicture}[
    scale = \myScale,
    ]

\tikzset{vertical label/.style={rotate=90,anchor=east}}   
\tikzset{diagonal label/.style={rotate=45,anchor=north east}}

\foreach \y in {1,...,\numClasses} 
{
    \node [anchor=east] at (0.4,-\y) {\pgfmathparse{\classNames[\y-1]}\pgfmathresult}; 
    
    \foreach \x in {1,...,\numClasses}  
    {
    \def\totSamples{0}
    \foreach \ll in {1,...,\numClasses}
    {
        \pgfmathparse{\myConfMat[\ll-1][\x-1]}   
        \xdef\totSamples{\totSamples+\pgfmathresult} 
    }
    \pgfmathparse{\totSamples} \xdef\totSamples{\pgfmathresult}  
    
    \begin{scope}[shift={(\x,-\y)}]
        \def\mVal{\myConfMat[\y-1][\x-1]} 
        \def\mValRatio{\myConfMatRatio[\y-1][\x-1]} 
        \pgfmathtruncatemacro{\r}{\mVal}   %
        \pgfmathtruncatemacro{\rRatio}{\mValRatio}   %
        \pgfmathtruncatemacro{\p}{round(\r/\totSamples*100)}
        \coordinate (C) at (0,0);
        \ifthenelse{\p<50}{\def\txtcol{black}}{\def\txtcol{white}} 
        \node[
            draw,                 
            text=\txtcol,         
            align=center,         
            fill=black!\p,        
            minimum size=\myScale*10mm,    
            inner sep=0,          
            ] (C) {\rRatio\\\p\%};     
        \ifthenelse{\y=\numClasses}{
        \node [] at ($(C)-(0,0.75)$) 
        {\pgfmathparse{\classNames[\x-1]}\pgfmathresult};}{}
    \end{scope}
    }
}
\coordinate (yaxis) at (-0.2,0.5-\numClasses/2);  
\coordinate (xaxis) at (0.5+\numClasses/2, -\numClasses-1.2); 
\node [vertical label] at (yaxis) {Predicted Class};
\node []               at (xaxis) {Target Class};
\end{tikzpicture}}}}

  \caption{Confusion matrices of the test set predictions by our best model for each subtask. The percentages show the ratio of the target class, which was predicted as that class. }
  \label{fig:confusion1}
\end{figure}
\section{Conclusion}

In this paper, we presented our approach for fact verification and evidence finding for tabular data in scientific documents. We show that transfer learning from \tabfact and standardization of the tables to have a single header helps improve our system's performance. We also show that having separate evidence finding models for entailed/refuted statements helps improve our system's performance in the second subtask. 

We also find that our model has a significant drop in performance on large tables since they are truncated to fit in the $512$ tokens, the maximum number of tokens supported by \tapas.

In future work, we would like to experiment with table pruning methods like Heuristic entity linking \citep{Chen2020TabFact:} or Heuristic exact match \citep{eisenschlos-etal-2020-understanding} so that the statement and table can fit in $512$ tokens. Our systems did not use the table metadata while making the predictions. In the future, we would also like to explore extending the model to encode table metadata along with the table.

\section*{Acknowledgments}

We thank the organisers of the shared task for their effort, and the anonymous reviewers for their insightful comments.

\bibliographystyle{acl_natbib}
\bibliography{anthology,acl2021}

\begin{thebibliography}{24}
\expandafter\ifx\csname natexlab\endcsname\relax\def\natexlab#1{#1}\fi

\bibitem[{Bowman et~al.(2015)Bowman, Angeli, Potts, and
  Manning}]{bowman-etal-2015-large}
Samuel~R. Bowman, Gabor Angeli, Christopher Potts, and Christopher~D. Manning.
  2015.
\newblock \href {https://doi.org/10.18653/v1/D15-1075} {A large annotated
  corpus for learning natural language inference}.
\newblock In \emph{Proceedings of the 2015 Conference on Empirical Methods in
  Natural Language Processing}, pages 632--642, Lisbon, Portugal. Association
  for Computational Linguistics.

\bibitem[{Chen et~al.(2020)Chen, Wang, Chen, Zhang, Wang, Li, Zhou, and
  Wang}]{Chen2020TabFact:}
Wenhu Chen, Hongmin Wang, Jianshu Chen, Yunkai Zhang, Hong Wang, Shiyang Li,
  Xiyou Zhou, and William~Yang Wang. 2020.
\newblock \href {https://openreview.net/forum?id=rkeJRhNYDH} {Tab{F}act: A
  large-scale dataset for table-based fact verification}.
\newblock In \emph{International Conference on Learning Representations}.

\bibitem[{Dagan et~al.(2006)Dagan, Glickman, and Magnini}]{10.1007/11736790_9}
Ido Dagan, Oren Glickman, and Bernardo Magnini. 2006.
\newblock The pascal recognising textual entailment challenge.
\newblock In \emph{Machine Learning Challenges. Evaluating Predictive
  Uncertainty, Visual Object Classification, and Recognising Tectual
  Entailment}, pages 177--190, Berlin, Heidelberg. Springer Berlin Heidelberg.

\bibitem[{Devlin et~al.(2019)Devlin, Chang, Lee, and
  Toutanova}]{devlin-etal-2019-bert}
Jacob Devlin, Ming-Wei Chang, Kenton Lee, and Kristina Toutanova. 2019.
\newblock \href {https://doi.org/10.18653/v1/N19-1423} {{BERT}: Pre-training of
  deep bidirectional transformers for language understanding}.
\newblock In \emph{Proceedings of the 2019 Conference of the North {A}merican
  Chapter of the Association for Computational Linguistics: Human Language
  Technologies, Volume 1 (Long and Short Papers)}, pages 4171--4186,
  Minneapolis, Minnesota. Association for Computational Linguistics.

\bibitem[{Eisenschlos et~al.(2020)Eisenschlos, Krichene, and
  M{\"u}ller}]{eisenschlos-etal-2020-understanding}
Julian Eisenschlos, Syrine Krichene, and Thomas M{\"u}ller. 2020.
\newblock \href {https://doi.org/10.18653/v1/2020.findings-emnlp.27}
  {Understanding tables with intermediate pre-training}.
\newblock In \emph{Findings of the Association for Computational Linguistics:
  EMNLP 2020}, pages 281--296, Online. Association for Computational
  Linguistics.

\bibitem[{Gupta et~al.(2020)Gupta, Mehta, Nokhiz, and
  Srikumar}]{gupta-etal-2020-infotabs}
Vivek Gupta, Maitrey Mehta, Pegah Nokhiz, and Vivek Srikumar. 2020.
\newblock \href {https://doi.org/10.18653/v1/2020.acl-main.210} {{INFOTABS}:
  Inference on tables as semi-structured data}.
\newblock In \emph{Proceedings of the 58th Annual Meeting of the Association
  for Computational Linguistics}, pages 2309--2324, Online. Association for
  Computational Linguistics.

\bibitem[{Herzig et~al.(2020)Herzig, Nowak, M{\"u}ller, Piccinno, and
  Eisenschlos}]{herzig-etal-2020-tapas}
Jonathan Herzig, Pawel~Krzysztof Nowak, Thomas M{\"u}ller, Francesco Piccinno,
  and Julian Eisenschlos. 2020.
\newblock \href {https://doi.org/10.18653/v1/2020.acl-main.398} {{T}a{P}as:
  Weakly supervised table parsing via pre-training}.
\newblock In \emph{Proceedings of the 58th Annual Meeting of the Association
  for Computational Linguistics}, pages 4320--4333, Online. Association for
  Computational Linguistics.

\bibitem[{Iyyer et~al.(2017)Iyyer, Yih, and Chang}]{iyyer-etal-2017-search}
Mohit Iyyer, Wen-tau Yih, and Ming-Wei Chang. 2017.
\newblock \href {https://doi.org/10.18653/v1/P17-1167} {Search-based neural
  structured learning for sequential question answering}.
\newblock In \emph{Proceedings of the 55th Annual Meeting of the Association
  for Computational Linguistics (Volume 1: Long Papers)}, pages 1821--1831,
  Vancouver, Canada. Association for Computational Linguistics.

\bibitem[{Liu et~al.(2019)Liu, Ott, Goyal, Du, Joshi, Chen, Levy, Lewis,
  Zettlemoyer, and Stoyanov}]{liu2019roberta}
Yinhan Liu, Myle Ott, Naman Goyal, Jingfei Du, Mandar Joshi, Danqi Chen, Omer
  Levy, Mike Lewis, Luke Zettlemoyer, and Veselin Stoyanov. 2019.
\newblock \href {http://arxiv.org/abs/1907.11692} {{RoBERTa}: A robustly
  optimized bert pretraining approach}.

\bibitem[{Neeraja et~al.(2021)Neeraja, Gupta, and
  Srikumar}]{neeraja-etal-2021-incorporating}
J.~Neeraja, Vivek Gupta, and Vivek Srikumar. 2021.
\newblock \href {https://doi.org/10.18653/v1/2021.naacl-main.224}
  {Incorporating external knowledge to enhance tabular reasoning}.
\newblock In \emph{Proceedings of the 2021 Conference of the North American
  Chapter of the Association for Computational Linguistics: Human Language
  Technologies}, pages 2799--2809, Online. Association for Computational
  Linguistics.

\bibitem[{Pasupat and Liang(2015)}]{pasupat-liang-2015-compositional}
Panupong Pasupat and Percy Liang. 2015.
\newblock \href {https://doi.org/10.3115/v1/P15-1142} {Compositional semantic
  parsing on semi-structured tables}.
\newblock In \emph{Proceedings of the 53rd Annual Meeting of the Association
  for Computational Linguistics and the 7th International Joint Conference on
  Natural Language Processing (Volume 1: Long Papers)}, pages 1470--1480,
  Beijing, China. Association for Computational Linguistics.

\bibitem[{Paszke et~al.(2019)Paszke, Gross, Massa, Lerer, Bradbury, Chanan,
  Killeen, Lin, Gimelshein, Antiga, Desmaison, Kopf, Yang, DeVito, Raison,
  Tejani, Chilamkurthy, Steiner, Fang, Bai, and Chintala}]{NEURIPS2019_9015}
Adam Paszke, Sam Gross, Francisco Massa, Adam Lerer, James Bradbury, Gregory
  Chanan, Trevor Killeen, Zeming Lin, Natalia Gimelshein, Luca Antiga, Alban
  Desmaison, Andreas Kopf, Edward Yang, Zachary DeVito, Martin Raison, Alykhan
  Tejani, Sasank Chilamkurthy, Benoit Steiner, Lu~Fang, Junjie Bai, and Soumith
  Chintala. 2019.
\newblock \href
  {http://papers.neurips.cc/paper/9015-pytorch-an-imperative-style-high-performance-deep-learning-library.pdf}
  {Py{T}orch: An imperative style, high-performance deep learning library}.
\newblock In H.~Wallach, H.~Larochelle, A.~Beygelzimer, F.~d\textquotesingle
  Alch\'{e}-Buc, E.~Fox, and R.~Garnett, editors, \emph{Advances in Neural
  Information Processing Systems 32}, pages 8024--8035. Curran Associates, Inc.

\bibitem[{Peters et~al.(2018)Peters, Neumann, Iyyer, Gardner, Clark, Lee, and
  Zettlemoyer}]{peters-etal-2018-deep}
Matthew Peters, Mark Neumann, Mohit Iyyer, Matt Gardner, Christopher Clark,
  Kenton Lee, and Luke Zettlemoyer. 2018.
\newblock \href {https://doi.org/10.18653/v1/N18-1202} {Deep contextualized
  word representations}.
\newblock In \emph{Proceedings of the 2018 Conference of the North {A}merican
  Chapter of the Association for Computational Linguistics: Human Language
  Technologies, Volume 1 (Long Papers)}, pages 2227--2237, New Orleans,
  Louisiana. Association for Computational Linguistics.

\bibitem[{Shi et~al.(2020)Shi, Zhang, Yin, and Liu}]{shi-etal-2020-learn}
Qi~Shi, Yu~Zhang, Qingyu Yin, and Ting Liu. 2020.
\newblock \href {https://www.aclweb.org/anthology/2020.coling-main.466} {Learn
  to combine linguistic and symbolic information for table-based fact
  verification}.
\newblock In \emph{Proceedings of the 28th International Conference on
  Computational Linguistics}, pages 5335--5346, Barcelona, Spain (Online).
  International Committee on Computational Linguistics.

\bibitem[{Suhr et~al.(2017)Suhr, Lewis, Yeh, and Artzi}]{suhr-etal-2017-corpus}
Alane Suhr, Mike Lewis, James Yeh, and Yoav Artzi. 2017.
\newblock \href {https://doi.org/10.18653/v1/P17-2034} {A corpus of natural
  language for visual reasoning}.
\newblock In \emph{Proceedings of the 55th Annual Meeting of the Association
  for Computational Linguistics (Volume 2: Short Papers)}, pages 217--223,
  Vancouver, Canada. Association for Computational Linguistics.

\bibitem[{Suhr et~al.(2019)Suhr, Zhou, Zhang, Zhang, Bai, and
  Artzi}]{suhr-etal-2019-corpus}
Alane Suhr, Stephanie Zhou, Ally Zhang, Iris Zhang, Huajun Bai, and Yoav Artzi.
  2019.
\newblock \href {https://doi.org/10.18653/v1/P19-1644} {A corpus for reasoning
  about natural language grounded in photographs}.
\newblock In \emph{Proceedings of the 57th Annual Meeting of the Association
  for Computational Linguistics}, pages 6418--6428, Florence, Italy.
  Association for Computational Linguistics.

\bibitem[{Thorne et~al.(2018)Thorne, Vlachos, Christodoulopoulos, and
  Mittal}]{thorne-etal-2018-fever}
James Thorne, Andreas Vlachos, Christos Christodoulopoulos, and Arpit Mittal.
  2018.
\newblock \href {https://doi.org/10.18653/v1/N18-1074} {{FEVER}: a large-scale
  dataset for fact extraction and {VER}ification}.
\newblock In \emph{Proceedings of the 2018 Conference of the North {A}merican
  Chapter of the Association for Computational Linguistics: Human Language
  Technologies, Volume 1 (Long Papers)}, pages 809--819, New Orleans,
  Louisiana. Association for Computational Linguistics.

\bibitem[{Wang et~al.(2021)Wang, Mahajan, Danilevsky, and
  Rosenthal}]{wang-etal-2021-semeval}
Nancy Xin~Ru Wang, Diwakar Mahajan, Marina Danilevsky, and Sara Rosenthal.
  2021.
\newblock {SemEval-2021 Task 9: A fact verification and evidence finding
  dataset for tabular data in scientific documents (SEM-TAB-FACTS)}.
\newblock In \emph{Proceedings of the 15th international workshop on semantic
  evaluation (SemEval-2021)}.

\bibitem[{Wolf et~al.(2020)Wolf, Debut, Sanh, Chaumond, Delangue, Moi, Cistac,
  Rault, Louf, Funtowicz, Davison, Shleifer, von Platen, Ma, Jernite, Plu, Xu,
  Le~Scao, Gugger, Drame, Lhoest, and Rush}]{wolf-etal-2020-transformers}
Thomas Wolf, Lysandre Debut, Victor Sanh, Julien Chaumond, Clement Delangue,
  Anthony Moi, Pierric Cistac, Tim Rault, Remi Louf, Morgan Funtowicz, Joe
  Davison, Sam Shleifer, Patrick von Platen, Clara Ma, Yacine Jernite, Julien
  Plu, Canwen Xu, Teven Le~Scao, Sylvain Gugger, Mariama Drame, Quentin Lhoest,
  and Alexander Rush. 2020.
\newblock \href {https://doi.org/10.18653/v1/2020.emnlp-demos.6} {Transformers:
  State-of-the-art natural language processing}.
\newblock In \emph{Proceedings of the 2020 Conference on Empirical Methods in
  Natural Language Processing: System Demonstrations}, pages 38--45, Online.
  Association for Computational Linguistics.

\bibitem[{Yang et~al.(2020)Yang, Nie, Feng, Liu, Chen, and
  Zhu}]{yang-etal-2020-program}
Xiaoyu Yang, Feng Nie, Yufei Feng, Quan Liu, Zhigang Chen, and Xiaodan Zhu.
  2020.
\newblock \href {https://doi.org/10.18653/v1/2020.emnlp-main.628} {Program
  enhanced fact verification with verbalization and graph attention network}.
\newblock In \emph{Proceedings of the 2020 Conference on Empirical Methods in
  Natural Language Processing (EMNLP)}, pages 7810--7825, Online. Association
  for Computational Linguistics.

\bibitem[{Yang et~al.(2019)Yang, Dai, Yang, Carbonell, Salakhutdinov, and
  Le}]{NEURIPS2019_dc6a7e65}
Zhilin Yang, Zihang Dai, Yiming Yang, Jaime Carbonell, Russ~R Salakhutdinov,
  and Quoc~V Le. 2019.
\newblock \href
  {https://proceedings.neurips.cc/paper/2019/file/dc6a7e655d7e5840e66733e9ee67cc69-Paper.pdf}
  {{XLNet}: Generalized autoregressive pretraining for language understanding}.
\newblock In \emph{Advances in Neural Information Processing Systems},
  volume~32. Curran Associates, Inc.

\bibitem[{Zhang et~al.(2020)Zhang, Wang, Wang, Cao, Zhang, and
  Wang}]{zhang-etal-2020-table}
Hongzhi Zhang, Yingyao Wang, Sirui Wang, Xuezhi Cao, Fuzheng Zhang, and
  Zhongyuan Wang. 2020.
\newblock \href {https://doi.org/10.18653/v1/2020.emnlp-main.126} {Table fact
  verification with structure-aware transformer}.
\newblock In \emph{Proceedings of the 2020 Conference on Empirical Methods in
  Natural Language Processing (EMNLP)}, pages 1624--1629, Online. Association
  for Computational Linguistics.

\bibitem[{Zhong et~al.(2017)Zhong, Xiong, and Socher}]{zhong2017seq2sql}
Victor Zhong, Caiming Xiong, and Richard Socher. 2017.
\newblock \href {http://arxiv.org/abs/1709.00103} {{Seq2SQL}: Generating
  structured queries from natural language using reinforcement learning}.

\bibitem[{Zhong et~al.(2020)Zhong, Tang, Feng, Duan, Zhou, Gong, Shou, Jiang,
  Wang, and Yin}]{zhong-etal-2020-logicalfactchecker}
Wanjun Zhong, Duyu Tang, Zhangyin Feng, Nan Duan, Ming Zhou, Ming Gong, Linjun
  Shou, Daxin Jiang, Jiahai Wang, and Jian Yin. 2020.
\newblock \href {https://doi.org/10.18653/v1/2020.acl-main.539}
  {{L}ogical{F}act{C}hecker: Leveraging logical operations for fact checking
  with graph module network}.
\newblock In \emph{Proceedings of the 58th Annual Meeting of the Association
  for Computational Linguistics}, pages 6053--6065, Online. Association for
  Computational Linguistics.

\end{thebibliography}


\end{document}